\documentclass[conference]{IEEEtran}
\IEEEoverridecommandlockouts
\usepackage{cite}
\usepackage{amsmath,amssymb,amsfonts}
\usepackage{algorithm}
\usepackage{algorithmic}
\usepackage{graphicx}
\usepackage{textcomp}
\usepackage{xcolor}
\usepackage{multirow}
\usepackage{subfigure}

\newcommand{\tco}{\textcolor{black}}
\def\BibTeX{{\rm B\kern-.05em{\sc i\kern-.025em b}\kern-.08em
    T\kern-.1667em\lower.7ex\hbox{E}\kern-.125emX}}
\begin{document}

\title{Learned Lossless JPEG Transcoding via Joint Lossy and Residual Compression}

\author{\IEEEauthorblockN{ Xiaoshuai Fan, Xin Li, Zhibo Chen\textsuperscript{*}\thanks{\textsuperscript{*}Corresponding author}}
\IEEEauthorblockA{\textit{Unviersity of Science and Technology of China, Hefei, China} \\
 \{fanxs, lixin666\}@mail.ustc.edu.cn, chenzhibo@ustc.edu.cn}
}

\newcommand{\ieno}{\textit{i.e.}}
\newcommand{\egno}{\textit{e.g.}}
\newcommand{\methodname}{TLRC}
\maketitle

\begin{abstract}
\tco{As a commonly-used image compression format, JPEG has been broadly applied in the transmission and storage of images. To further reduce the compression cost while maintaining the quality of JPEG images, lossless transcoding technology has been proposed to recompress the compressed JPEG image in the DCT domain. Previous works, on the other hand, typically reduce the redundancy of DCT coefficients and optimize the probability prediction of entropy coding in a hand-crafted manner that lacks generalization ability and flexibility. To tackle the above challenge, we propose the learned lossless JPEG transcoding framework via \textbf{J}oint \textbf{L}ossy and \textbf{R}esidual \textbf{C}ompression. Instead of directly optimizing the entropy estimation, we focus on the redundancy that exists in the DCT coefficients. To the best of our knowledge, we are the first to utilize the learned end-to-end lossy transform coding to reduce the redundancy of  DCT coefficients in a compact representational domain. We also introduce residual compression for lossless transcoding, which adaptively learns the distribution of residual DCT coefficients before compressing them using context-based entropy coding. Our proposed transcoding architecture shows significant superiority in the compression of JPEG images thanks to the collaboration of learned lossy transform coding and residual entropy coding. Extensive experiments on multiple datasets have demonstrated that our proposed framework can achieve about 21.49\% bits saving in average based on JPEG compression, which outperforms the typical lossless transcoding framework JPEG-XL by 3.51\%.}
\end{abstract}


\section{Introduction}
\tco{With the development of image acquisition and processing, daily production of billions of images places a heavy burden on storage and transmission capacity. To tackle the above challenge, a number of standard image compression approaches such as JPEG~\cite{JPEG1}, JPEG2000~\cite{JPEG2000}, JPEG XR~\cite{JPEGXR}, as well as learned image codecs~\cite{learned_lossy_codec1,learned_lossy_codec2,learned_lossy_codec3,learned_lossy_codec10,learned_lossy_codec11, lossless2,lossless3,lossless4,lossless5,lossless6, guo20203, wu2021learned} and related optimization strategies~\cite{liu2022swiniqa,li2020multi,li2022hst}, have been proposed in order to reduce the compression cost while maintaining the image quality. Among them, JPEG compression format is a simple and flexible format used in real life, which has led to the majority of images being stored in this format. However, the JPEG codec is insufficiently optimized for rate-distortion performance, resulting in significant redundancy in the bitstreams. To further reduce compression costs, it is necessary and urgent to identify a suitable framework for recompressing JPEG images. }
\begin{figure*}
\centering 
\includegraphics[width=0.7\textwidth]{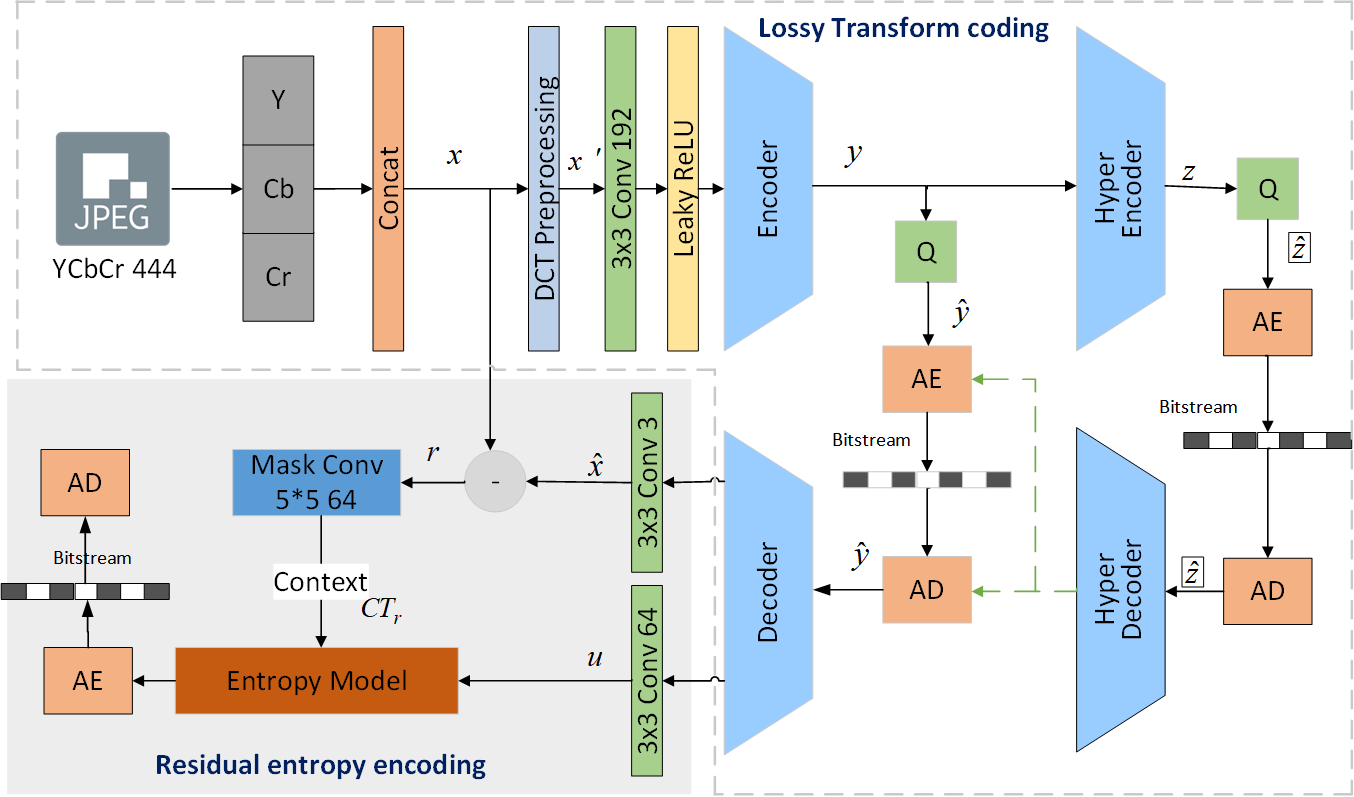}
\caption{Overall architecture of the proposed Learned Lossless JPEG Transcoding method. It consists of Lossy Transform coding and Residual entropy coding. AE and AD stand for arithmetic coding and arithmetic decoding respectively. Q stands for quantization module. The residual calculation is element-wise subtraction. $\hat{y}$, $\hat{z}$ and $r$ need to be encoded as the bitstreams.}
\label{fig:Network}
\vspace{-3mm}
\end{figure*}

\tco{As the commonly-used compression method, JPEG codec\cite{JPEG1}} follows the block-based compression approach. It divides the input image into $8\times 8$ pixel blocks, transforms each block to a DCT domain, and then encodes the DC and AC coefficients using entropy coding. \tco{To reduce spatial redundancy between adjacent blocks, the JPEG codec utilizes the DC coefficients of the reference block to predict the coding block and compresses only the residual components. However, since the simple transform and prediction modes, there is still huge redundancy in bit streams, which wastes an enormous amount of storage and transmission bandwidth. Lossless JPEG transcoding has been proposed to recompress JPEG images in the DCT domain in order to further reduce compression costs while maintaining the qualities of the JPEG format. Traditional lossless transcoding works can be roughly divided into three categories, including 1) optimizing the prediction of DCT coefficients~\cite{Trad-1,Trad-2,Trad-3,Trad-5}, 2) optimizing the entropy coding model~\cite{Trad-6,Trad-7,Trad-8,Trad-9,Trad-12}, and 3) prepossessing the DCT coefficients for more efficient entropy coding~\cite{Trad-10,Trad-11}.}
There are also standard software packages, such as Lepton~\cite{Lepton} and JPEG XL~\cite{JPEGXL1,JPEGXL2}, that perform well with lossless JPEG transcoding. Lepton is primarily concerned with optimizing entropy models and symbol representations. JPEG XL attains a higher compression ratio by extending the $8\times 8$ DCT to a variable-size DCT, allowing the block size to be 8, 16 or 32. Moreover, JPEG XL substitutes Asymmetric Numeral Systems for Huffman coding. However, the above methods heavily rely on hand-crafted mode design, which makes it hard to eliminate redundant data in an effective and efficient manner.

\tco{To tackle the above challenge, in this paper, we propose a learned end-to-end lossless JPEG transcoding framework to further reduce the redundancy of JPEG images in an adaptive and sufficient manner. Specifically, we aim to implement lossless transcoding from two perspectives, \ieno, how to reduce spatial redundancy in DCT coefficients and how to accurately estimate the probability distribution. In contrast to traditional JPEG transcoding methods~\cite{Trad-1,Trad-2,Trad-3,Trad-5,Trad-6,Trad-7,Trad-8,Trad-9,Trad-12,Trad-10,Trad-11}, which only process redundancy in the DCT domain, we use commonly-used transform coding in learned lossy compression~\cite{vae1} to learn an optimal compact representation as~\cite{DCTimage} for the compression of DCT coefficients based on the amount of of nature images. Specifically, we rearrange the JPEG quantized DCT coefficients into the DCT image based on the different frequency components, which makes it easier for the network to capture the spatial redundancy. Then, for lossy compression of DCT coefficients, we employ the typical end-to-end transform coding backbone~\cite{learned_lossy_codec1}.} 

\tco{Thanks to the transform coding, there is little correlation in the residual of DCT coefficients. Therefore, we adopt learned entropy coding, which is capable of learning a relatively precise probability distribution for each residual coefficient. We extract adaptively the side information for probability estimation from the decoded DCT coefficients and their residual. Since there is sparsity, we model the residual probability using the mixture logistic probability model. With the cooperation of learned lossy transform coding and residual entropy coding, an advanced end-to-end lossless JPEG transcoding framework can be constructed. Extensive experiments on multiple benchmarks, including DIV2K~\cite{DIV2k}, Kodak~\cite{Kodak}, CLIC validation~\cite{CLIC} have demonstrated the effectiveness of our proposed method.} 



\tco{The main contributions of our work can be summarized as follows:
\begin{itemize}
    \item \tco{We propose a learned end-to-end lossless JPEG transcoding framework by solving the challenges from two perspectives, \ieno, how to reduce the spatial redundancy existing in the DCT coefficients, and how to estimate the probability accurately for entropy coding.} 
    \item \tco{We introduce the first framework for lossless JPEG transcoding with joint lossy compression and residual entropy coding.}
    \item \tco{Extensive experiments on multiple benchmarks have demonstrated the effectiveness and superiority of our proposed JPEG transcoding framework.}
\end{itemize}}

\tco{The rest of the paper is organized as follows: In Sec.~\ref{sec: method}, We will describe in detail our framework for learning lossless JPEG transcoding from the perspective of principles and implementations. Sec.~\ref{sec: experiments} describes our experimental setting and results to validate the effectiveness of our scheme. Finally, we conclude this paper in Section~\ref{sec: conclusion}.}

\section{Method}
\label{sec: method}
\subsection{Overview of Learned Lossless JPEG Transcoding Framework}
\tco{The lossless JPEG transcoding aims to further reduce the compression cost by recompressing the JPEG images in the DCT domain. That means compressing the quantified DCT coefficients $x$ with a lossless optimization goal as Eq.~\ref{eq:MSE}. }
\begin{equation}
    \centering
    min ~J=~R+\lambda D(x, \Tilde{x}), where~D(x,\Tilde{x}) = (x-\Tilde{x})^2 = 0,
    \label{eq:MSE}
\end{equation}

\tco{where $\Tilde{x} $ are the lossless transcoding reconstructed coefficients. R, D and J are respectively as the rate, distortion, and rate-distortion performance. We analyze the essential challenges in lossless transcoding and then propose a framework \methodname~for learned lossless JPEG transcoding by addressing two typical challenges, \ieno, how to reduce the spatial redundancy existing in DCT coefficients and how to accurately estimate the probability distribution. Our proposed \methodname~is shown in Fig.~\ref{fig:Network}, which consists of two crucial components, \ieno, lossy transform coding module and residual entropy module. Among them, lossy transform coding is utilized to determine the optimal representation space for removing spatial redundancy. For precise entropy coding, the residual entropy module employs a learnable probability distribution prediction. Combining lossy transform coding and residual entropy coding allows for the generation of more compact bit streams for the compression of qualified DCT coefficients. In the sections that follow, we will discuss in detail the preprocessing of qualified DCT coefficients, lossy transform coding, and residual entropy coding. }

\subsection{Preprocessing of Quantified DCT Coefficients}
\label{subsec: DCT image}
\begin{figure}[t]
    \centering
    \includegraphics[width=1.0\linewidth]{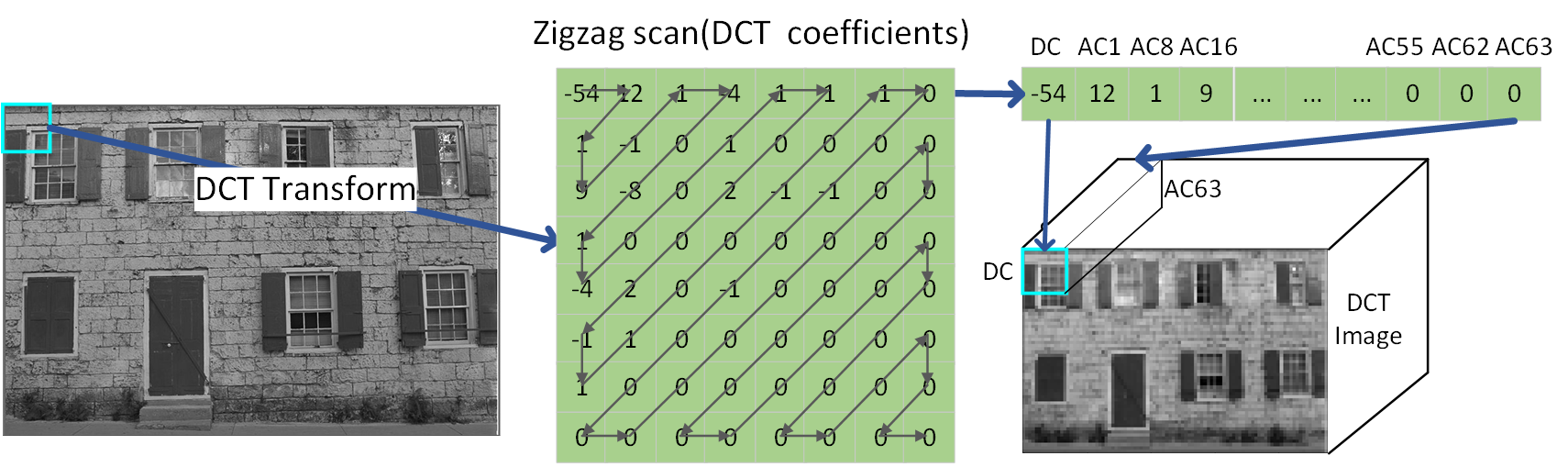}
    \caption{Preprocessing of Quantified DCT Coefficients. Each $8 \times 8$ block is flattened using the indicated zigzag scanning, and the collection of quantized DCT vectors can be reformed into a DCT image, with $8\times8=64$ channels.}
    \label{fig:DCT image}
    \vspace{-6mm}
\end{figure}

\tco{The quantified DCT coefficients of each block in the JPEG codec are stored in a $8 \times 8$ matrix as shown in Fig.~\ref{fig:DCT image}, which is hard to compress directly due to the interference of coefficients of different frequencies in the same spatial space. To transform the DCT coefficients into a format that can be effectively compressed, we follow~\cite{DCTimage} and preprocess the quantified DCT coefficients into frequency-based DCT images.} 
\tco{As shown in Fig.~\ref{fig:DCT image}, there are 64 coefficients corresponding to 64 frequency components for each block. We first utilise Zig-Zag scan to reshape the coefficients into one-hot vectors ascending in frequency from low to high. The spatial dimension is then constructed by combining the coefficients of the same frequency component. Finally, we can obtain a DCT image with the shape $H\times W \times C$, where $C$ is 64 and $H$ and $W$ are the DCT image's height and width, respectively. The DCT images are suitable for compression because 1) the components from the same frequency are aggregated in the same spatial dimension, which simplifies the spatial redundancy elimination of lossy transform coding. 2) The frequency components are consistent in the dimension of the channel, allowing for more efficient entropy coding. }


\subsection{Lossy transform coding for DCT image}
\tco{After obtaining the processed DCT image $x'$ with the shape of $H \times W \times C$, we employ learned lossy transform coding to determine the optimal compact representation $y$ for the end-to-end compression of the DCT image. The components of learned lossy transform coding are non-linear transformation, relaxed quantization, and entropy coding. In particular, non-linear transform aims to transform the input DCT image into a representation of features that is compact and easily compressed. Relaxed quantization is used to simulate the process of hard quantization and ensure the backward propagation of the gradient. Entropy coding objectives for estimating rates during optimization and converting quantified coefficients to bit steam. As shown in in Fig.~\ref{fig:Network}, as our lossy transform coding module, we employ the commonly-used hyper-prior based compression model in~\cite{learned_lossy_codec1}. Given one preprocessed DCT image $x'$, the lossy transform coding module transforms DCT image $x'$ to a compact representation $y$ with the non-linear transform, which consists of a series of fundamental modules, convolution, leaky relu~\cite{leakyrelu}, and GDN. The representation $y$ is then quantified to a group of discrete coefficients $hat{y}$ and transformed to a bit stream that used entropy coding. To improve the accuracy of entropy coding, the lossy transform coding module transmits to the decoder the side information $hat{z}$ for distribution estimation using the hyper-prior module. Therefore, the rate $R {(\hat{y},\hat{z})}$ of lossy DCT image compression can be represented as Eq.~\ref{eq:r}:
\begin{equation}
    \centering
   R_{(\hat{y},\hat{z})}=\mathbb{E}_{p(x')}\mathbb{E}_{q_{\phi}(\hat{y},\hat{z}|x')}[-log(p_{\theta}(\hat{y}|\hat{z}))-log(p_{\theta}(\hat{z})],
    \label{eq:r}
\end{equation}
where $p(x')$ is the probability distribution of x', $p_\theta$ refers to the probability distribution given the parameter $\theta$ and $q_{\phi}(\hat{y},\hat{z}|x')$ is the prior probability of $(\hat{y}, \hat{z})$, which is modeled with Gaussian distribution. }

\subsection{Residual Entropy coding}
\begin{figure}[t]
    \centering
    \includegraphics[width=0.9\linewidth]{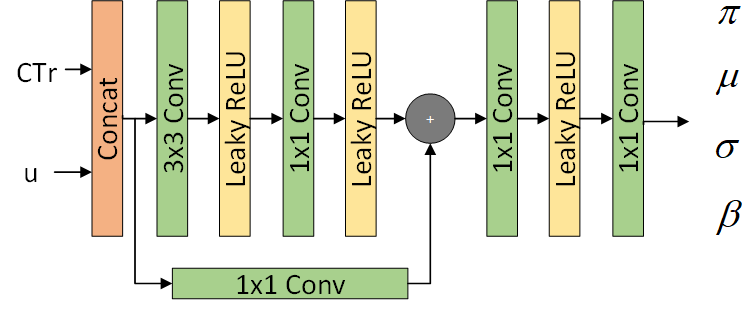}
    \caption{Entropy parameter estimation sub-network. Given $u$ and $CT_{r}$, estimates parameters of discrete logistic mixture likelihoods. All 1 × 1 convolutional layers except the last layer have 128 channels. The last convolutional layer has $3 \cdot K$ channels. K stands for the number of mixture logistic distribution. }
    \label{fig:enmtropy}
    \vspace{-4mm}
\end{figure}

\tco{To implement lossless JPEG transcoding, we calculate the residual $r$ between the original DCT coefficients $x$ and the decoded lossy DCT coefficients $\hat{x}$, and then adapt autoregressive context module based entropy coding to the residual $r$. As shown in Fig.~\ref{fig:Network},   context information $CT_r$ is extracted from the residual $r$ using the mask convolution~\cite{maskconv}. Then, we concatenate the context information $CT_r$ and the reconstructed DCT information $u$ and estimate the probability estimation for logistic entropy coding with the module shown in Fig.~\ref{fig:enmtropy}. This module contains both simple convolution and leaky relu layers.} 
\tco{In particular, we represent the residual coefficients in the  $i^{th}$ spatial location of Y, Cr, Cb components as $r_{i, Y}$, $r_{i, Cr}$, and $r_{i, Cb}$. The distribution of residual $r$ can be represented as Eq.~\ref{eq:PMF}:}
\begin{equation}
    \centering
    \begin{aligned}
    p_{\theta}(r|u,CT_{r})=\prod_{i}p_{\theta}(r_{i})|u_{i},CT_{r_{i}}),
    \end{aligned}
    \label{eq:PMF}
\end{equation}
where the distribution of the $i^{th}$ residual $p_{\theta}(r_{i})$ is computed with Eq.~\ref{eq:PMF2} by introducing the prediction between different color components.
\begin{gather}
    p_{\theta}(r_{i}|u,CT_{r}) = p_{\theta}(r_{i,Y}|u_{i},CT_{r_{i}}) \cdot 
    p_{\theta}(r_{i,Cr})|r_{i,Y},u_{i},CT_{r_{i}}) \notag \\ \cdot p_{\theta}(r_{i,Cb})|r_{i,Y},r_{i,Cr},u_{i},CT_{r_{i}}) 
    \label{eq:PMF2}
\end{gather}
\tco{We further model the probability mass function (PMF) of the residual $r$ with discrete logistic mixture likelihood~\cite{pixelcnn++}. As shown in Fig.~\ref{fig:enmtropy}, we introduce a Entropy Model to estimate entropy parameters, including mixture weights $\pi_{i}^k$, means $\mu_{i}^k$, variances $\sigma_{i}^k$ and mixture coefficients $\beta_{i}^k$. $k$ denotes the index of the k-th logistic distribution. The channel autoregression over $r_{i,Y},r_{i,Cr},r_{i,Cb}$ is implemented by updating the means:
\begin{gather}
    \mu_{i,Y}^k = \mu_{i,Y}^k , \mu_{i,Cr}^k = \mu_{i,Cr}^k + \beta_{i,Y} \cdot r_{i,Y}, \notag\\ 
    \mu_{i,Cb}^k = \mu_{i,Cb}^k + \beta_{i,Cr} \cdot r_{i,Y} + \beta_{i,Cb} \cdot r_{i,Cr}, \\
    p_{\theta}(r_{i}|u_{i},CT_{r}) \sim \sum_{k}\pi_{i}^k \cdot logistic(\mu_{i}^k,\sigma_{i}^k) \notag
\end{gather} 
}

\tco{Depart from the constrain on the rate terms $ R_{\hat{y},\hat{z}} $ and $R_{r}$, we add a distortion term $ D_{lossy}(x,\hat{x})=(x - \hat{x})^2 $ as the loss term to minimize the mean square error (MSE) between input x and lossy reconstruction $\hat{x}$. The full loss function for Learned JPEG image Transcoding is
\begin{equation}
    \centering
    Loss = R_{\hat{y},\hat{z}} + R_{r} + \lambda \cdot D_{lossy},
    \label{eq:loss}
\end{equation}}
where $\lambda$ controls the “rate-distortion” trade-off. We choose $\lambda$ as 0.03 empirically in this paper.

\section{Experiments}
\label{sec: experiments}
\subsection{Experimental Setup}
\label{subsec: experiments setup}
\tco{\noindent\textbf{Dataset:} Our training dataset consists of 10,000 JPEG compressed images selected randomly from the ImageNet~\cite{ILSVRC15} training set.  We evaluate our model on four popular datasets for compression:  Kodak~\cite{Kodak} dataset with 24 images, DIV2K~\cite{DIV2k} validation set with 100 images, CLIC~\cite{CLIC} professional validation dataset with 41 images and CLIC mobile validation dataset with 61 images. Since our work is designed for lossless JPEG transcoding, the test datasets have been compressed by the JPEG codec to extract the quantified DCT coefficients. }

\tco{\noindent\textbf{Implementation Details:} We first train only the lossy DCT image compression framework, where the rate is modeled by the entropy of quantized latent variables and the reconstruction distortion is measured by PSNR~\cite{PSNR}. Then we load the pretrained lossy compression model and jointly train Lossy transform coding and Residual entropy to achieve end to end Lossless JPEG Transcoding. And our Lossless JPEG Transcoding is optimized for 500 epochs using Adam~\cite{da2014method} with minibatches of size 16. The learning rate is set to $1 \times {10}^{-4} $ for the first 450 epochs and decays to $1 \times  {10}^{-5} $ for the remaining 50 epochs.}

\subsection{Performance}

\tco{\textbf{Performance comparison with the state-of-the-arts.} To validate the effectiveness of our proposed \methodname, we compare it with multiple commonly-used lossless JPEG transcoding methods including Lepton~\cite{Lepton} and JPEG XL~\cite{JPEG1} on four datasets. As shown in Table~\ref{tab:performance}, our \methodname~ achieves obvious bits saving of 21.49\%, which outperforms the Lepton by 1.97\% in average.}  

\begin{table}[H]
    \centering
    \vspace{-3mm}
    \caption{Performance comparison on various datasets}
    \setlength{\tabcolsep}{0.1mm}{
    \begin{tabular}{c|llll}
    
        \hline
         & \multicolumn{3}{c}{BPP and Bit Saving(\%)}\\
        \hline
        Method & Kodak & DIV2K & CLIC.mobile & CLIC.pro \\
        \hline
        JPEG~\cite{JPEG1} & 3.392 & 3.219 & 2.833 & 2.882\\
        \hline
        Lepton~\cite{Lepton}  & 2.777 \textcolor{green}{18.12\%} & 2.564 \textcolor{green}{20.35\%} &  2.281 \textcolor{green}{19.48\%} & 2.302 \textcolor{green}{20.13\%}\\
        \hline
        JPEG XL~\cite{JPEGXL1}  & 2.782 \textcolor{green}{17
        98\%} & 2.670 \textcolor{green}{17
        04\%} &  2.339 \textcolor{green}{17.44\%} & 2.321 \textcolor{green}{19.47\%}\\
        \hline
        Ours &  2.665 \textbf{\textcolor{green}{21.43\%}} &  2.526 \textbf{\textcolor{green}{21.53\%}} & 2.225 \textbf{\textcolor{green}{21.46\%}} & 2.261 \textbf{\textcolor{green}{21.55\%}} \\
        \hline
    \end{tabular}}
    \label{tab:performance}
    \vspace{-3mm}
\end{table}

\tco{We also analyze the proportion of each bit stream component. The bit stream file is divisible into two components: the lossy bit stream file and the residual bit stream file. As shown in TABLE.~\ref{tab:performance2},  the bit stream for residual accounts for nearly 80\% of the total bit rate, whereas lossy bit streams only account for about 20\% bit rate, indicating that the entropy coding portion of the residual can be further explored and optimized. }

\begin{table}[H]
    \centering
    \vspace{-3mm}
    \caption{Lossy part and residual bitstreams analysis}
    \begin{tabular}{c|cccc}
        \hline
        Item & Kodak & DIV2K & CLIC.mobile & CLIC.pro \\
        \hline
        lossy bin & 0.513 & 0.477 & 0.449 & 0.465 \\
        \hline
        residual bin &  2.152 & 2.049 & 1.776& 1.796\\
        \hline
        total & 2.665 & 2.526 & 2.225 & 2.261 \\
        \hline
        res / total & 80.75\% & 81.16\% & 79.82\% & 79.43\% \\
        \hline
    \end{tabular}
    \label{tab:performance2}
    \vspace{-3mm}
\end{table}

\tco{\textbf{Performance on different quality levels.} \tco{We investigate the generalization capability of our proposed \methodname by directly applying the model trained on JPEG quality level 95 to different JPEG quality levels, including 55, 65, 75 and 85 in Fig.~\ref{fig:QP}. Our method's superiority is demonstrated by the fact that \methodname can be adapted well to different JPEG quality levels, as shown in the figure. }
\begin{figure}[t]
    \centering
    \includegraphics[width=0.85\linewidth]{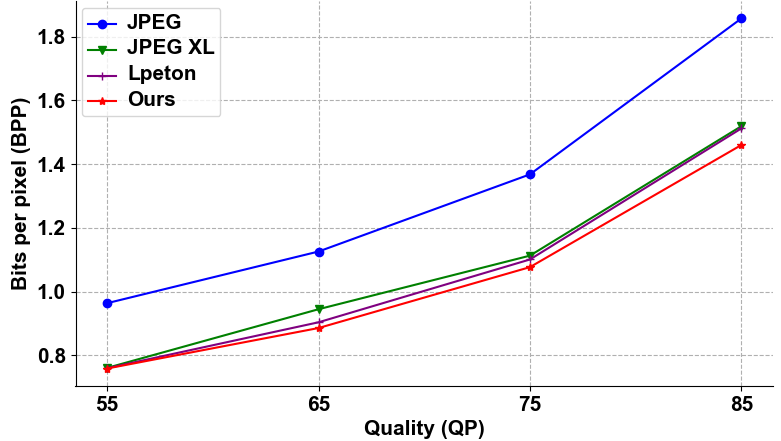}
    \caption{Comparison of bits per pixel (BPP) on Kodak dataset when recompressing JPEG images of different quality levels (QP = 55, 65, 75, 85). We use the model trained with QP 95}
    \label{fig:QP}
    \vspace{-2mm}
\end{figure}}

\subsection{Ablation Study}
\tco{To verify the effectiveness of our lossy-plus-residual framework, we remove the bit stream compressed by lossy DCT and store the extracted feature information $u$ as the side information for the entropy encoding model. Then, the Laplace distribution is implemented to directly encode the entropy of the original DCT image. As demonstrated in Tab.~\ref{tab:performance3}, removing the lossy transform coding module will result in an average performance decrease of 5.02\%, indicating that transform coding is crucial for the compression of DCT coefficients. }
\begin{table}[H]
    \centering
    \vspace{-3mm}
    \caption{Effectiveness of Lossy DCT image compression}
    \setlength{\tabcolsep}{0.1mm}{
    \begin{tabular}{c|llll}
        \hline
         & \multicolumn{3}{c}{BPP and Bit Saving(\%)}\\
        \hline
        Method & Kodak & DIV2K & CLIC.mobile & CLIC.pro \\
        \hline
        JPEG~\cite{JPEG1} & 3.392 & 3.219 & 2.833 & 2.882 \\
        \hline
        ours(w/o lossy) &  2.834 \textcolor{green}{16.45\%} & 2.683 \textcolor{green}{16.65\%} & 2.372 \textcolor{green}{16.27\%} & 2.406 \textcolor{green}{16.52\%}  \\
        \hline
        Ours &  2.665 \textbf{\textcolor{green}{21.43\%}} &  2.526 \textbf{\textcolor{green}{21.53\%}} & 2.225 \textbf{\textcolor{green}{21.46\%}} & 2.261 \textbf{\textcolor{green}{21.55\%}} \\
        \hline
    \end{tabular}}
    \label{tab:performance3}
    \vspace{-3mm}
\end{table}

\section{Conclusion}
\tco{In this paper, we analyze the essential challenges in JPEG lossless transcoding, including 1) how to reduce spatial redundancy in DCT coefficients and 2) how to accurately estimate the probability for entropy coding. To address the mentioned issues, we propose an innovative lossless JPEG transcoding framework \methodname~ that combines lossy transform coding and residual entropy coding. Lossy transform coding aims to reduce spatial redundancy in a compact form learned representation space. To learn an accurate probability distribution for entropy coding, residual entropy coding is employed. Extensive experiments and ablation studies on four representative datasets have demonstrated the effectiveness of our proposed \methodname.}
\label{sec: conclusion}

\section*{Acknowledge}
This work was supported in part by NSFC under Grant U1908209, 62021001 and
the National Key Research and Development Program of China 2018AAA0101400

\bibliographystyle{IEEEtran}
\bibliography{TLRC}
\end{document}